\pdfoutput=1

\def\year{2019}\relax
\documentclass[letterpaper]{article} 
\usepackage{aaai19}  
\usepackage{times}  
\usepackage{helvet}  
\usepackage{courier}  
\usepackage{url}  
\usepackage{graphicx}  
\nocopyright

\usepackage{amsmath}
\usepackage{amssymb}
\usepackage{amsfonts}

\begin{document}
%
\title{Responsive Planning and Recognition for Closed-Loop Interaction$^{1}$}
\author{Richard G. Freedman \\ Smart Information Flow Technologies, LLC \\ \\rfreedman@sift.net 
\And Yi Ren Fung \and Roman Ganchin \and Shlomo Zilberstein \\ University of Massachusetts Amherst\\
 College of Information and Computer Sciences\\
\{freedman, yfung, rganchin, shlomo\}@cs.umass.edu\\
}
\maketitle
\begin{abstract}
Many intelligent systems currently interact with others using at least one of fixed communication inputs or preset responses, resulting in rigid interaction experiences and extensive efforts developing a variety of scenarios for the system.  Fixed inputs limit the natural behavior of the user in order to effectively communicate, and preset responses prevent the system from adapting to the current situation unless it was specifically implemented.  Closed-loop interaction instead focuses on dynamic responses that account for what the user is currently doing based on interpretations of their perceived activity.  Agents employing closed-loop interaction can also monitor their interactions to ensure that the user responds as expected.  We introduce a closed-loop interactive agent framework 
that integrates planning and recognition 
 to predict what the user is trying to accomplish and autonomously decide on actions to take in response to these predictions.  Based on a recent demonstration\footnote{This paper is an extension of a previously accepted extended abstract for the demo track at ICAPS 2019.  The title and authors are the same. Sections~\ref{sec:intro},~\ref{sec:pretcil}, and \ref{sec:conclusion} are modified.  Section~\ref{sec:challenges} is original and based on the demonstration of the previously accepted extended abstract's work.\label{revisionDisclaimer}} of such an assistive 
interactive agent in a turn-based simulated game, we also discuss new research challenges that are not present in the areas of artificial intelligence planning or recognition alone.
\end{abstract}

\section{Introduction\label{sec:intro}}
From entertainment to personal assistance, intelligent systems are interacting with people in a variety of applications.  However, even when these systems appear to act autonomously and allow the user free will, there is usually extensive back-end development to engineer the interactive experience.  Though not as restrictive as expert systems with hand-coded tables of what to exactly do in every considerable situation, there is usually a fixed set of inputs or outputs that is mapped from or to artificial intelligence algorithms.  For example, natural language interfaces might perform speech-to-text and then map that text to a set of expected inputs through parsing or machine learning.  Likewise, embodied agents might have a preprogrammed finite state machine that specifies what output behavior to perform, and task and motion planning algorithms determine how to execute those behaviors given the current environment's configuration.

Even though these intelligent systems exhibit artificial intelligence and account for the environment and stimuli, they are not actually interacting with an understanding of the user.  People act with purpose, explore their environment, make mistakes, and will sometimes change their mind in the middle of doing something.  \textit{In the diverse domains for service robots, this will especially apply as people perform a large variety of tasks amongst their daily life routines}.  A robot that is cleaning should consider what activities people are doing nearby---then it can avoid introducing noise when others are listening to something or having a conversation, reduce interrupting activities to tidy up an area where people will be/are, and conveniently be ready to handle a mess shortly after it is made.  Likewise, robots that assist skilled trade workers, such as chefs, construction workers, plumbers, and mechanics, can prepare spaces and tools, as well as perform complementary tasks, after simply observing how human workers start to address their assigned task(s).  Closed-loop interaction addresses this by \textit{modeling users and making decisions with respect to those models}.

\begin{figure}
\centering
\includegraphics[scale=0.3]{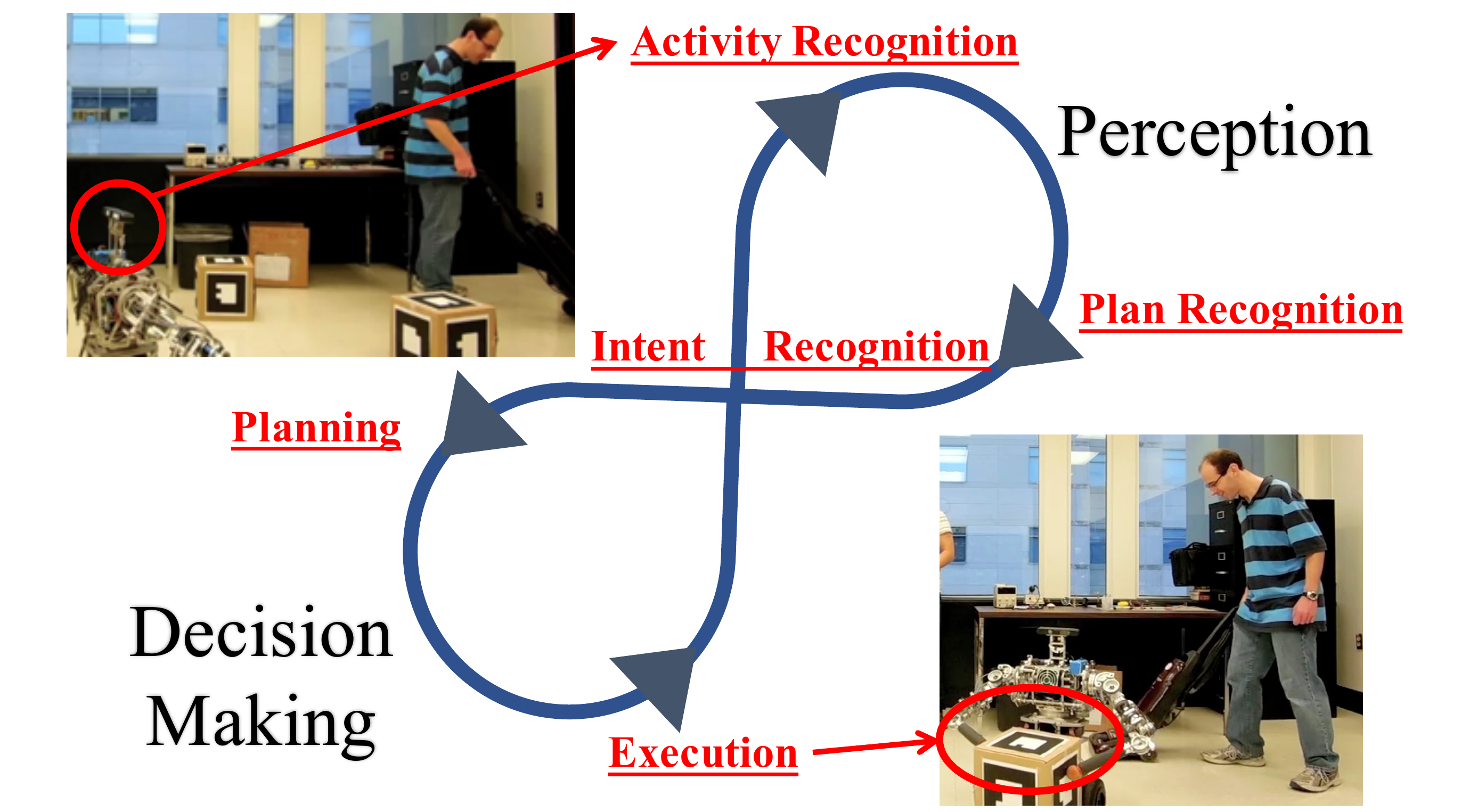}
\caption{The {\sc PReTCIL} framework's general flow for how perception and decision making affect each other in closed-loop interaction.\label{fig:pretcil}}
\end{figure}

We thus introduce the Planning and Recognition Together Close the Interaction Loop ({\sc PReTCIL}) framework as a cognitive architecture, which is illustrated in Figure~\ref{fig:pretcil}.  Similar frameworks that integrate planning and recognition for closed-loop interaction either rely on a library of precomputed plans for less robust recognition and simpler execution monitoring \cite{levine_ICAPS147945,jair_levineWilliams}
or require negotiation with the user to confirm the agent's understandings and actions \cite{geibPetrick_2016}.  Instead, {\sc PReTCIL} iterates indefinitely to \textit{update the recognized intents and plans using its perception and expected responses of the user} while also \textit{revising its decisions of how to act based on these updates}.

As a general framework, any appropriate algorithms can be applied to {\sc PReTCIL}.  For this paper, we implemented {\sc PReTCIL} using responsive planning \cite{fz_aaai2017} and recognition as planning \cite{DBLP:conf/aaai/RamirezG10} as its primary components.  Following an overview of these components and how they play a role in {\sc PReTCIL}, we discuss new research challenges that arose during a recent demonstration of the implementation.  These challenges explore novel problems that the traditional artificial intelligence methods did not face prior to both integration with other algorithms and use in human interactions.

\section{{\sc PReTCIL} Overview and Implementation\label{sec:pretcil}}
Illustrated in Figure~\ref{fig:pretcil}, the primary feature of {\sc PReTCIL} is the \textit{integration of perceiving} interaction partners \textit{and making decisions} about how to respond such that these two aspects influence each other throughout the interactive experience.  Activity recognition abstracts low-level observations, such as raw sensor information, into higher-level action labels/descriptions that can be used for plan recognition and intent recognition.  These two forms of recognition estimate the interaction partner's goals and predict what actions they will take to achieve them.  With these predictions, a planner can select its own actions to respond to the estimated goals with respect to the expected actions.  If the decisions are too high-level, then execution will determine how the agent can perform the chosen actions.  Given this response, intent recognition can also predict how the interaction partner will respond to the agent's action.  Lastly, activity recognition completes the interaction loop by again abstracting what the interaction partner does to confirm whether or not the predicted response is correct.

\subsection{Perception via Planning as Recognition\label{sec:pretcil.perception}}
When the interactive experience begins, the assistive agent has no model of the interactive partners.  This means that the agent is not aware of what they want to do and must first observe them in order to make any informed decisions.  For the demonstration, users played a turn-based game in a simulated environment with the freedom to select from a set of completion criteria---successfully satisfying any one of them resulted in winning the game.  The user performed an action on the first turn, which provides some information to the agent running our implementation of the {\sc PReTCIL} framework about which criteria they intended to complete.  Then the agent performed an action if it received sufficient information to decide how to respond. 

Due to the simulated game setting of this demonstration, user inputs were limited to discrete key/mouse 
 presses that are easily identifiable without any raw sensor data.  Thus this implementation of {\sc PReTCIL}
 simply performs activity recognition as a mapping from the input to the game's corresponding action.

The plan and intent recognition components receive these actions as observations for probabilistic recognition as planning \cite{DBLP:conf/aaai/RamirezG10}.  This class of algorithms runs a generative planner to simulate the user solving a variety of problems and then compares their solutions in order to identify which of the completion criteria are most likely.  The key assumption applied in recognition as planning is that the user \textit{is acting as optimally as possible to achieve their goal}.  This means that 
the observed actions either lead to completing the criteria (optimal to perform) or are out of the criteria's way (not optimal to perform).  The more likely criteria will have a greater difference between these solutions. 

The plans generated for these comparisons serve as the output for plan recognition, providing information about what the user is expected to do by themself when satisfying each completion criteria.  The distribution over the different criteria, computed using the costs of these recognized plans, is the output for intent recognition because it identifies how likely each criteria is motivating the user's actions.

\subsection{Decision Making via Responsive Planning\label{sec:pretcil.dec_making}}
When deciding how to respond to the interactive partner's possible intents, it is important to consider the long-term interaction as much as the current action being taken.  This is especially important at the beginning of the interactive experience because their initial actions are often relevant to completing multiple criteria---this ambiguity is the worst-case distinctiveness \cite{keren_ICAPS147814} that measures the maximum number of actions that can be shared between the start of two optimal plans solving different goals from the same initial state.  
Furthermore, assisting the interactive partner towards the completion criteria that they did not select can hinder the experience and reduce their trust and willingness to work with the assistive agent.

Our implementation of the {\sc PReTCIL} framework accounts for this by identifying the necessities \cite{fz_aaai2017}, which are shared features between the goals that the assistive agent believes the user is most likely completing.  With respect to the distribution over the possible criteria that the intent recognition component provides, this is the weighted sum over the parts of each completion criteria.  It is rarely the case that different intents are mutually exclusive of each other; so the agent can assist the user by completing the common tasks that progress towards all the likely completion criteria until the user performs some action that further disambiguates their intent.

The necessities generate an intermediate goal for the agent, and the planner used for probabilistic recognition as planning in Section~\ref{sec:pretcil.perception} can find a sequence of actions that will accomplish this generated goal.  The planner generates the plan from the current state and assigns actions to both the agent and the user each turn until the intermediate goal is accomplished.  This joint solution is not revealed to the user in the demonstration, but is necessary for the assistive agent to realize how the state might change after the user's turns.  While this plan is active, the agent executes the next action in response to the user.  Like with activity recognition, this demonstration's simulated game setting allows us to simply map the action to animation and update the state.

\subsection{Execution Monitoring\label{sec:pretcil.exec_monitor}}
Although the joint plan derived in Section~\ref{sec:pretcil.dec_making} is not conveyed to the user in the demonstration, the actions assigned to the user are assumed to take place in order for the agent's future actions to execute successfully without uncertainty.  Our implementation of the {\sc PReTCIL} framework thus uses this plan for the second purpose of intent recognition to predict how the user will respond to the agent's actions each turn.  If the user's action returned from the activity recognition component matches, then we assume that the interaction is going smoothly and execute the agent's next action in the joint plan.  If the user's action does not match, then there is a chance that the agent recognized incorrectly and reassesses the 
 completion criteria with the newest observation.  This execution monitoring system completes the interaction loop.

\section{Challenges for Closed-Loop Interaction \label{sec:challenges}}
Our implementation of the {\sc PReTCIL} framework described above was demonstrated at the Twenty-Ninth International Conference on Planning and Scheduling (ICAPS) in July 2019.  Approximately fifteen conference attendees watched others interact or directly interacted with the demonstration over the one-hour period that it was on display.  Based on the authors' observations of their experiences and feedback, new questions and research challenges emerged involving the integration of the two areas within an interactive domain.

Although we have IRB approval to run interactive sessions with human subjects for experimental purposes, this demonstration was not related to any experiments. 
 We 
 only discuss general observations and experiences to avoid risk of revealing any participant identities.

\subsection{Brief Explanation of the Demo\label{sec:challenges.demo}}
The demonstration environment is based on a common toy problem in the planning community called Block Words (also called Blocks World).  A table contains stacks of blocks that each have a letter inscribed on them, and the agent is tasked with picking-and-placing blocks until there exists a stack of blocks whose inscribed letters spell a specific goal word when read from top-to-bottom.  Actions either pick up a block that is on top of any stack or put down a held block on top of any stack, and an agent can only hold up to one block at a time.  Our extension for assitive interaction includes two agents, each able to hold up to one block at a time, that take turns performing actions.  Either agent may pass their turn with a no-op action.

\begin{figure}
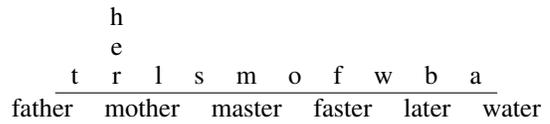

\centering
\begin{tabular}{cccccccccc}
& h & & & & & & & &  \\
& e & & & & & & & &  \\
t & r & l & s & m & o & f & w & b & a \\
\hline
\end{tabular}
\begin{tabular}{cccccc}
father & mother & master & faster & later & water \\
\end{tabular}
\caption{The initial block layout (above) and possible goal words to spell (below) in the demonstration.\label{fig:blockwordsDemo}}
\end{figure}

The initial block layout and possible words to spell are illustrated in Figure~\ref{fig:blockwordsDemo}, based on the setup 
 used in comparison experiments with \citeauthor{thesis_levine}'s \shortcite{thesis_levine} interactive agent. 
 For the purposes of our demonstration, the user always made the first move after announcing the specific word they wanted to spell. 
 The user and agent took turns until the user was done (whether or not the goal was accomplished).  The demonstration was reset between users. 
 
\subsection{Challenge Topics\label{sec:challenges.topics}}
\paragraph{Sufficient Information to Interact} For the demonstration, two parameter values were set manually when initiating an interactive session: the necessities threshold and the number of turns that the user has for a head-start.  The former adjusts the sensitivity of feature selection when generating intermediate goals from the recognized distributions; the threshold $\tau \in [0, 1]$ requires a feature to appear in enough goal criteria that they collectively represent at least $\tau$ of the distribution.  The latter acts as a delay before the assistive agent begins responding, 
 which allows it to have an observation sequence that is less ambiguous during recognition.  Although parameter tuning is a common challenge in many algorithms, especially for machine learning performance, we have identified some impacts for choosing different values.

If the necessities threshold is too low, then more features unique to specific goal criteria are added to the intermediate goal.  Although this sounds more robust to accommodate the uncertainty at the beginning of the interaction, the present-day norm of \textit{conjunctive goal conditions} means that there is a greater opportunity for the goal to have contradictions.  In our demonstration, at most one block can be placed on top of another.  However, lower thresholds allowed words that shared one letter to require both of their preceding letters on top---``mother'' and ``father'' could easily require both the `a' and `o' blocks to be placed on top of the `t' block (in addition to building the stack that spells ``ther'') when $\tau$ is sufficiently small. 
 The lack of a solution to this goal means that the agent will not be able to find a plan and act that turn, which made it appear less helpful to the user.  Likewise, if the necessities threshold is too high, then no features might be found and the intermediate goal is to change nothing---``mother'' and ``father'' may have a combined probability of $0.8$ in some cases, but that does not identify any intermediate goal conditions when $\tau = 0.9$. 
 In this case, the \textit{solution of doing nothing has the same consequence as not finding a solution to a goal with contradicting conditions}.

The number of head-start turns can be more drastic.  If it is low, such as $0$ to begin interacting immediately, then the handful of observations can be very ambiguous such that the assistive agent recognizes a near-uniform distribution over the subset of goal criteria that use those actions at least once in their possible solutions.  Even with a reasonable necessities threshold, this distribution can either be spread too thin to find no features for the intermediate goal or be concentrated enough over almost-distinct goal criteria that contradicting unique features are added to the intermediate goal.  The latter scenario sometimes selected unique features that did not contradict others, but were generally incorrect so that the agent performed actions that did not make sense to the users.  An additional observation would have often pruned those goal criteria from the recognition algorithm, which is why we added the head-start parameter.  When it was set too high, though, the user made enough progress that they found the agent's late response less useful. 



\vspace{-12pt}
\paragraph{(Un)Intentional Communication} When the autonomous agent in human-robot interactions has its own personal goals, it can communicate its intentions towards these goals to the human via legible planning with low-level motions \cite{Dragan-2013-7732} or high-level actions \cite{DBLP:conf/aaai/KulkarniSK19}.  However, our assistive agent's personal goal is more abstract: ``to help the user with their own goal.''  So the agent does not have a personal goal until an intermediate one is recognized and computed.  \citeauthor{geibPetrick_2016} \shortcite{geibPetrick_2016} account for communicating the agent's newfound goals during the negotiation step, but their assistive agent pipelines the interaction process so that no further recognition is performed after negotiating its goals. 
 We assumed that the \textit{cognitive load of frequent negotiations would not be ideal} as {\sc PReTCIL} loops indefinitely.

However, our implementation's planner assumes what the user will do, which is sometimes reordered or extraneous enough to confuse the user instead.  In one instance, the user looked at the debug data to read the assumed actions and mentioned that providing this expectation would have been a useful explanation for the unexpected behavior.  Providing explanations for decision making systems \cite{xaip} has been growing in popularity recently, but we need to be careful that these explanations do not constrain the user's freedom to act in accordance to what the machine does \cite{DBLP:conf/aaaifs/CastroRMB17}.

Some users at the demonstration already succumbed to such constraints when selecting their own actions to ensure legibility to the recognition algorithm, viewing the demonstration as a puzzle rather than an open-ended interactive experience.  Does this \textit{defeat the purpose of closed-loop interaction} if people adjust their own behaviors \textit{to satisfy the algorithms around them rather than act naturally}?  Though we mentioned that \citeauthor{jair_levineWilliams}'s \shortcite{jair_levineWilliams} assistive agents have more restricted interactions using 
 a library of precomputed plans, this library often contains multiple plans that allow flexibility to the interactive partner (this is the purpose behind their \textit{choice nodes} where the human can take one of several actions).  This leads to a research challenge for finding the balance in a hybrid of closed-loop interaction frameworks.  If a joint-agent planner finds multiple plans to the intermediate goal, then which plans' action should be used when there are multiple matches to the next observation?  That is, when monitoring the execution, which plans are ``going according to plan''?




\vspace{-13.5pt}
\paragraph{What Information Actually Matters?} While most of the challenges discussed so far involve general issues that relate specifically to the interactive experience, it is also important to consider some algorithmic challenges. 
 The most critical ones we identified during the demonstration relate to \textit{using all the available information}.  Some plan recognition algorithms already address noisy sensing \cite{DBLP:conf/ijcai/SohrabiRU16} and irrelevant experimental actions while exploring the environment \cite{DBLP:journals/tist/MirskyGS17}, but these methods still assume that the observed agent is the only actor in the world.  Planning algorithms can address various forms of uncertainty, but we are unaware of any that consider the uncertainty of the goal's validity.

The assistive agent's actions also change the world, and these need to be acknowledged during recognition.  We simply encoded them as observations because recognition as planning handles missing observations by assuming actions 
 that can connect two consecutive observations 
 were performed.  However, the potential for poorly chosen intermediate goals 
 threw off the recognition algorithms due to the agent's sometimes incorrect actions and state modifications.  Accounting for them as noise or experimentation might work pragmatically, but they are conceptually different because these actions have purpose and influence the interactive partner's later actions toward their goal.  Furthermore, for long-term interactive systems that cannot be reset like our demonstration, how should observation sequences be modified over time for relevancy to the current interaction only?

When our demonstration's assistive agent computes a joint plan with its intermediate goal, it currently uses the same search heuristics; the state space and set of actions change to address turn-taking.  However, the above issues with useless goals present two things to consider.  First, when the goal contains contradicting conditions, is there a way to find a plan that satisfies some largest possible subset of conditions so that the agent can do something?  Second, if the agent is unable to find a plan, should the agent perform a default action or replan for some default goal?  We programmed our assistive agent to perform a no-op, but this led to a few failed demonstrations where the user needed a block that the agent was holding before it failed to find a plan.  Even if these users intended to confuse the assistive agent with noisy observations, a default goal of 
 not holding any blocks would at least allow the user to complete the task on their own. 





\section{Conclusion\label{sec:conclusion}}
For less structured interactions between users and intelligent

\noindent systems, closed-loop interaction that perceives what people do and decides how to appropriately respond is necessary.  We introduced the {\sc PReTCIL} framework as a cognitive architecture for such interaction and implemented it as an assistive agent for a game.  A recent demonstration 
 revealed new research challenges for artificial intelligence methods involved in closed-loop interaction. 
 Future research will explore these challenges, but we encourage the artificial intelligence for human-robot interaction community to consider their own solutions and identify additional problems.  \textit{Many novel situations from interactive experiences and integrated frameworks take these traditional algorithms out of their original context}, and we need to address them as we continue to study and create intelligent interactive systems.
 
 \section*{Acknowledgements} Richard G. Freedman thanks Steven Levine for his frequently engaging discussions regarding artificial intelligence for closed-loop interaction.  The authors also thank the anonymous reviewers for their feedback.  A final thank you goes to the ICAPS 2019 demonstration participants. \newpage 

\bibliography{NSF-PR,freedmanThesis}
\bibliographystyle{aaai}

\end{document}